%% file: main.tex
\documentclass[lettersize,journal]{IEEEtran}
\usepackage{amsmath,amsfonts}
\usepackage{algorithmic}
\usepackage{algorithm}
\usepackage{array}
\usepackage[caption=false,font=normalsize,labelfont=sf,textfont=sf]{subfig}
\usepackage{textcomp}
\usepackage{stfloats}
\usepackage{url}
\usepackage{verbatim}
\usepackage{graphicx}
\usepackage{cite}

\usepackage{epsfig}
\usepackage{graphicx}
\usepackage{amsmath}
\usepackage{amssymb}
\usepackage{booktabs}
\usepackage{multirow}
\usepackage{bbm}
\usepackage{soul}
\usepackage{xcolor,colortbl}
\definecolor{ffe1da}{RGB}{255,225,218}
\definecolor{F7E0D5}{RGB}{247,224,213}
\definecolor{darkF7E0D5}{RGB}{209,154,128}

\begin{document}

\title{ClickVOS: Click Video Object Segmentation}
\author{
Pinxue Guo, Lingyi Hong, Xinyu Zhou, Shuyong Gao, Wanyun Li, Jinglun Li, \\ Zhaoyu Chen, Xiaoqiang Li, Wei Zhang, Wenqiang Zhang
\thanks{Pinxue Guo, Jinglun Li, and Zhaoyu Chen are with the Shanghai Engineering Research Center of AI\&Robotics, Academy for Engineering\&Technology, Fudan University, Shanghai, China (e-mail: pxguo21@m.fudan.edu.cn; jingli960423@gmail.com; haoyuchen20@fudan.edu.cn).}
\thanks{Lingyi Hong, Xinyu Zhou, Shuyong Gao, Wanyun Li, and Wei Zhang are with the Shanghai Key Lab of Intelligent Information Processing, School of Computer Science, Fudan University, Shanghai, China (e-mail: lyhong22@m.fudan.edu.cn; zhouxinyu20@fudan.edu.cn; sygao18@fudan.edu.cn; wyli22@m.fudan.edu.cn; weizh@fudan.edu.cn).}
\thanks{Xiaoqiang Li is with the School of Computer Engineering and Science, Shanghai University, Shanghai, China (e-mail: xqli@shu.edu.cn).}
\thanks{Wenqiang Zhang is with Engineering Research Center of AI\&Robotics, Ministry of Education, Academy for Engineering\&Technology, Fudan University, Shanghai, China, and also with the Shanghai Key Lab of Intelligent Information Processing, School of Computer Science, Fudan University, Shanghai, China (e-mail: wqzhang@fudan.edu.cn).}
\thanks{(Corresponding author: Wei Zhang and Wenqiang Zhang.)}
}

\markboth{Journal of \LaTeX\ Class Files,~Vol.~14, No.~8, August~2021}%
{Shell \MakeLowercase{\textit{et al.}}: A Sample Article Using IEEEtran.cls for IEEE Journals}

\IEEEpubid{0000--0000/00\$00.00~\copyright~2021 IEEE}

\maketitle

\begin{abstract}
Video Object Segmentation~(VOS) task aims to segment objects in videos. 
However, previous settings either require time-consuming manual masks of target objects at the first frame during inference or lack the flexibility to specify arbitrary objects of interest. 
To address these limitations, we propose the setting named Click Video Object Segmentation (ClickVOS) which segments objects of interest across the whole video according to a single click per object in the first frame.
And we provide the extended datasets DAVIS-P and YouTubeVOS-P that with point annotations to support this task.
ClickVOS is of significant practical applications and research implications due to its only 1-2 seconds interaction time for indicating an object, comparing annotating the mask of an object needs several minutes.
However, ClickVOS also presents increased challenges. To address this task, we propose an end-to-end baseline approach named called Attention Before Segmentation (ABS), motivated by the attention process of humans. ABS utilizes the given point in the first frame to perceive the target object through a concise yet effective segmentation attention. Although the initial object mask is possibly inaccurate, in our ABS, as the video goes on, the initially imprecise object mask can self-heal instead of deteriorating due to error accumulation, which is attributed to our designed improvement memory that continuously records stable global object memory and updates detailed dense memory.
In addition, we conduct various baseline explorations utilizing off-the-shelf algorithms from related fields, which could provide insights for the further exploration of ClickVOS. The experimental results demonstrate the superiority of the proposed ABS approach.
Extended datasets and codes will be available at https://github.com/PinxueGuo/ClickVOS.
\end{abstract}


\input{./sections/introduction.tex}
\input{./sections/related_work.tex}

\input{./sections/task.tex}

\input{./sections/method.tex}
\input{./sections/baseline.tex}
\input{./sections/experiments.tex}
\input{./sections/conclusion.tex}

\section{Acknowledgments}
This work was supported by National Natural Science Foundation of China (No.62072112), Scientific and Technological innovation action plan of  Shanghai Science and Technology Committee (No.22511102202, No.22511101502, and No.21DZ2203300).

\bibliographystyle{IEEEtran}
\bibliography{ref}

\end{document}

%% file: sections/introduction.tex
\section{Introduction}
\IEEEpubidadjcol 

\IEEEPARstart{V}{ideo} Object Segmentation (VOS)~\cite{liu2020guided, caelles2017one, fan2021semi, gao2023decoupling, bhat2020learning, tang2023holistic, robinson2020learning, li2018video, perazzi2017learning, hu2018videomatch, hong2021adaptive, oh2019stm, yang2020collaborative, guo2022adaptive, hong2023lvos} is a fundamental task in computer vision, which aims to segment target objects in a video sequence. It has a wide range of applications, including video editing~\cite{oh2018fast, wang2017copy, liu2021temporal}, surveillance~\cite{zhao2016real,patil2021unified, ahmed2018trajectory, liu2019perceiving}, robotics~\cite{Griffin_2020_WACV}, and 
autonomous driving~\cite{kim2020video, yuan2021temporal, zhang2018end}. 

VOS can be mainly categorized into two subtasks based on whether providing the initial mask of the object in the first frame during inference.
Semi-supervised Video Object Segmentation (SemiVOS) algorithms can segment arbitrary interested objects at the instance level with the mask annotation for the first frame. But the precise interaction time of several minutes severely limits the application scenarios of SemiVOS. 
Unsupervised Video Object Segmentation~(UnVOS) algorithms, although do not require manual interaction, cannot specify the objects to be segmented.
Instead, they automatically segment objects from the background at the object level, which simply predicts all salient objects in the video together as a single binary mask.

\begin{figure}[t]
    \centering
    \includegraphics[width=1.0\linewidth]{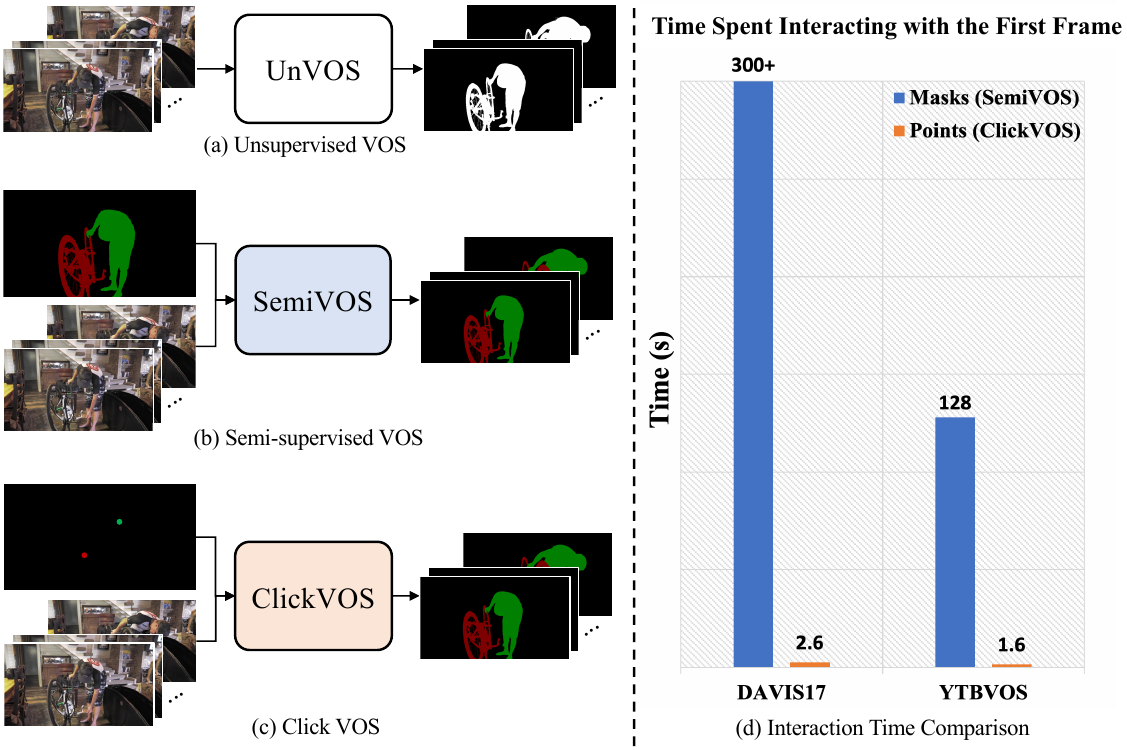}
    \caption{Different VOS tasks and the interaction time comparison between SemiVOS and ClickVOS. SemiVOS requires time-consuming manual masks of target objects at the first frame during inference while UnVOS lack the flexibility to specify arbitrary objects of interest.}
    \label{fig_motivation}
\end{figure}

\IEEEpubidadjcol 
\textit{How can we both indicate arbitrary objects of interest and avoid using the time-consuming and labour-intensive masks?}
The simplest way to indicate a target object is a single click on it. 
However, there is a blank slate in the research of using only one point to segment each object in the video. 
To address this gap, we propose to tackle the novel task named Click Video Object Segmentation~(ClickVOS), which aims at segmenting target objects across the whole video according to a single click per object in the first frame. 
And to support this task, we provide the DAVIS16-P, DAVIS17-P, and YouTubeVOS-P datasets by extending the widely-used datasets DAVIS-2016\cite{davis16}, DAVIS-2017\cite{davis17}, and YouTubeVOS\cite{ytvos18}.
ClickVOS is of significant practical applications and research implications. As illustrated in Fig.~\ref{fig_motivation}, annotating objects in the first frame of an inference video with masks is a time-consuming process~\cite{bearman2016s}, taking an average of 128 seconds on the YouTubeVOS~\cite{ytvos18} dataset and over 300 seconds on the DAVIS~\cite{davis17} dataset. In contrast, using points for annotation takes only 1.6 and 2.6 seconds, respectively, according to our statistics. This will unlock numerous applications that involve video object segmentation, such as autonomous driving, mobile applications, and other user-facing scenes where either the available interaction time is limited or the environment is not conducive to precise interaction.

While ClickVOS shows significant potential, it also poses greater challenges. Firstly, the click point only provides an approximate position of the object. Obtaining an accurate segmentation mask with only the position in a single frame remains a challenging problem. Secondly, without the precise mask of the first frame, all previous approaches of relying on the first frame as reliable information in SemiVOS are no longer feasible.

To tackle the ClickVOS task with these challenges, we take inspiration from human attention and propose an end-to-end approach, Attention Before Segmentation~(ABS).
From the perspective of human attention in handling the specific object segmentation task in a video, humans do not necessarily require pixel-level dense annotations as in SemiVOS. Instead, we can quickly perceive the approximate target with the single indication as the visual system of Gestalt principles~\cite{wertheimer1923untersuchungen}. Then, by continuously attending to the object and getting further detailed information, such as the object edge from the motion cue, the video object segmentation process task is completed. 
Our ABS performs almost the same process, starting with rough attention and then performing precise segmentation, closely following the process described above. According to the given points in the first frame, object tokens with identity embedding for each object are encoded with the Point Tokenizer. These object tokens estimate the masks in the first frame through the Segment Attention, where the masks could be imprecise. Then as the video goes on, the initially imprecise object mask can self-heal instead of deteriorating due to error accumulation suffered in prior VOS methods, This is attributed to our designed improvement memory, which continuously records stable global object memory and updates detailed dense memory.

In addition, we conduct several baseline explorations using off-the-shelf algorithms from related fields, such as SemiVOS~\cite{oh2019stm, yang2022decoupling}, UnVOS~\cite{ventura2019rvos}, interactive VOS~\cite{mivos}, instance segmentation~\cite{maskrcnn}, and interactive image segmentation~\cite{fbrs2020,kirillov2023segment}. These efforts not only provide experience for other researchers to tackle the new task but also demonstrates the superiority of the ABS approach we proposed.

Our contributions in this paper are summarized below:
\begin{itemize}
    \item We tackle the novel task, named Click Video Object Segmentation~(ClickVOS), which aims at segmenting objects of interest in all video frames, given a single point per object in the first frame.
    \item To support this task, we provide the extended datasets DAVIS-P and YouTubeVOS-P enabling further exploration of this new task.
    \item We propose the end-to-end baseline approach, called Attention Before Segmentation~(ABS), which simulates the attention process of humans, possesses self-healing capabilities, and performs well in ClickVOS task.
    \item We conduct various baseline explorations using off-the-shelf algorithms from related fields, providing invaluable insights to future researchers and demonstrating the superiority of our proposed ABS approach.
\end{itemize}

%% file: sections/related_work.tex
\section{Related Work}

\subsection{Semi-supervised Video Object Segmentation}
Exploiting the given first frame masks, which indicate target objects, and segmenting them across the whole video is the essential problem of SemiVOS~\cite{zhu2021separable, fan2021semi, liu2020guided, li2018video, caelles2017one, hu2018videomatch, guo2022adaptive}. 
Recently, considerable studies have made significant efforts in this area.
These studies can be categorized into three distinct paradigms: propagation-based method, model-based method and matching-based method. 
Propagation-based methods~\cite{li2018video, perazzi2017learning} trained an object mask propagator to adjust a misaligned mask to align with the target object in the current frame. MaskTrack \cite{perazzi2017learning} operates on a frame-by-frame basis, utilizing the mask from previous frame to guide it towards the object of interest in the succeeding frame. DyeNet \cite{li2018video} introduce a re-identification module and an attention-based mask propagation method, which aim to recover missing objects and improve performance.
But the implicit target-appearance modelling strategy is prone to lead to error accumulation. 
Model-based methods~\cite{caelles2017one, bhat2020learning, robinson2020learning} online fine-tuning a segmentation model to adapt to various objects. The target appearance model of FRTM~\cite{robinson2020learning} comprises a lightweight module that is learned using fast optimization techniques. This module is used to predict a coarse egmentation result. LWL \cite{bhat2020learning} proposes a differentiable few-shot learner, which is designed to predict a target model for imporving the segmentation accuracy.
Matching-based methods~\cite{hu2018videomatch, voigtlaender2019feelvos, oh2019stm, yang2020collaborative, guo2022adaptive, hong2021adaptive}, the currently prevailing methodology, conduct semantic matching mechanism at the pixel level, which computig the simliarity between previous and current frames. STMVOS~\cite{oh2019stm} proposes a memory network and aims to read relevant feature from previous frames relying on the matching mechanism. CFBI \cite{yang2020collaborative} designs a foreground-background collaborative matching mechanism to reduce interference from similar objects.
Despite their success, the time-consuming mask annotation required in the first frame severely limits the application scenarios of SemiVOS.

\subsection{Unsupervised Video Object Segmentation}
UnVOS \cite{luo2017unsupervised, li2018instance, wang2019zero, li2018unsupervised, dutt2017fusionseg, tokmakov2017learning, cheng2017segflow, tokmakov2019learning, zhou2020motion, hong2023simulflow} perform object segmentation without manual object masks during inference.
UnVOS performs object segmentation without manual object masks during inference.
Pixel instance embedding methods~\cite{li2018instance, wang2019zero} produce the pixel-wise feature embeddings and then compute the similarity between these embeddings, determining whether the pixels belong to the same object. End-to-end models have become mainstream in UnVOS, which can be categorized into two paradigms: Short-term information encoding methods~\cite{dutt2017fusionseg, cheng2017segflow, li2018unsupervised, tokmakov2019learning, zhou2020motion} exploit recurrent neural networks or two-stream networks to combine the spatial and temporal information. And long-term information encoding methods~\cite{lu2019see, zhang2020unsupervised, ren2021reciprocal} usually construct a siamese network to extract features in frame pairs and caculate the feature correlation to utilize the global temporal information. 
Although UnVOS algorithms can segment salient foreground objects, they lack the flexibility to select specific targets. Whereas ClickVOS is able to segment arbitrary objects according to the points indicated.

\subsection{Interactive Video Object Segmentation}
Besides ClickVOS, there is another setting also utilizing points, Interactive Video object Segmentation (InterVOS), where humans draw scribbles~\cite{oh2019fast, miao2020memory, heo2020interactive, yin2021learning} or clicks~\cite{chen2018blazingly, mivos} on a selected frame and algorithms compute the segmentation maps for all video frames. This process is repeated continuously for refinement. Most methods are achieved by separating the InteractiveVOS into an interactive module and a propagation module. The interactive module outputs the segmentation and refines the prediction after user interaction. And the propagation module propagates the refined masks to other frames across the video. This task concerns the efficiency of human interaction time but needs a human in the loop to produce results and is mainly used for data annotation. In contrast, ClickVOS does not require continuous human interaction, and each video only needs to be clicked once in the first frame, which can decouple the human from the loop and has more significant practical applications.

%% file: sections/task.tex
\section{ClickVOS}

In this section, we give the formal definition of the ClickVOS task (Sec.~\ref{problem define}) and then introduce the datasets and evaluation metrics of this task (Sec.~\ref{Metrics}). Moreover, we discuss some details and challenges of the task (Sec.~\ref{discussion}).

\subsection{Problem Definition}
\label{problem define}

The click video object segmentation task aims at segmenting objects of interest in all video frames, given a single point per object in the first frame. To formulate, given a video with $T$ frames  $ \{I_t \in \mathbb{R}^{3\times H\times W}\}_{t=1}^{T} $ and a point per object in the first frame $ P_1 = \{ p_1^n \}_{n=0}^N $ for indicating the $N$ interested targets, the goal of ClickVOS is to predict the object masks in all frames $ \{ M_t \in \mathbb{N}^{H\times W} \}_{t=1}^T $. Notably, there is one and only one point for each object, and $ p_1^0 $ is the point for the background. To clarify, the above description refers to the inference process of the ClickVOS task, and the mask annotation as the ground truth is available during the training stage.

\subsection{Evaluation}
\label{Metrics}

\subsubsection{Evaluation Datasets}
To explore the ClickVOS, we provide the DAVIS16-P, DAVIS17-P and YouTubeVOS-P benchmark datasets by extending the DAVIS \cite{davis16, davis17} and YouTubeVOS \cite{ytvos18} datasets in video object segmentation. We hope they can serve the community and inspire the development of the ClickVOS algorithm.
DAVIS benchmarks are widely used in video object segmentation with high-quality mask annotation. DAVIS16 is a single-object benchmark dataset containing 30/20 videos in the training/validation set. DAVIS17 is a multi-object benchmark whose training/validation set contains 60/30 videos.
YouTubeVOS is a large-scale dataset that contains 6459/1063 unique object instances in 3471/507 videos for training/validation set. Among them are 26 unseen categories that have not been seen in training data and 65 seen categories. 

First, we extend the validation set of the above datasets by annotating each object with a point at the first frame of the videos. Three annotators participate in this annotation effort, and for each image, we randomly select one of the annotations labeled by the three annotators to reduce personal bias.
Then for the training set, all frames are annotated with points by automatically selecting point annotations from the existing mask annotations. Specifically, the mask is first corroded for each foreground object, and then a random position of the mask is selected as the point annotation. 
On the one hand, automatically selecting point annotations from existing mask annotations can significantly save labor and time costs. On the other hand, manually annotated points are typically concentrated in the center of the object due to human attention, potentially biasing the model to make better predictions based on these "good" points that provide additional prior information during inference. Therefore, training with point annotations with some randomness can lead to a more robust model. 

\subsubsection{Evaluation Metrics}
We use the evaluation metrics from the DAVIS benchmark~\cite{davis16}: $ \mathcal{J} $, $ \mathcal{F} $, and $\mathcal{J\&F}$. The region accuracy $ \mathcal{J} $ calculates the average intersection-over-union (IoU) of the estimated mask and the ground truth mask. The boundary accuracy $ \mathcal{F} $ calculates the average boundary similarity between the prediction and the ground truth. And  $\mathcal{J\&F}$ denotes the average of $ \mathcal{J} $ and $ \mathcal{F} $.

\subsection{Discussion}
\label{discussion}

\subsubsection{Novel Challenges of ClickVOS}
Despite the practical significance shown by ClickVOS, it is considered to be a more challenging task. 
In contrast to SemiVOS, which starts with a known precise mask leading most advanced approaches to rely on dense matching with the precise reference mask, the ClickVOS task poses a different challenge. In ClickVOS, the first frame only provides points indicating the object rather than a precise mask, making it challenging as the appearance of the object to be segmented is unknown. This complicates the design of models, as we cannot rely on the premise that the object mask in the first frame is accurate.
Compared to UnVOS, which is now predominantly trained to segment only the most salient object in a video, the ClickVOS task involves segmenting arbitrary objects based on custom requirements. This poses a challenge for unsupervised models in terms of focusing on and perceiving the object categories they can handle.

\subsubsection{Target Object in ClickVOS}
With the advancement of interactive segmentation methods, there has been a rise in efforts focusing on segmenting objects at different levels of granularity, such as segmentation for whole or part. It is noted that ClickVOS primarily concerns the whole object granularity. For instance, if a click is placed on a person, the segmentation should encompass the entire person rather than specific body parts like the head or limbs. This maintains consistency with the VOS task and the associated dataset, enabling a fair and convenient exploration of the effect of substituting mask interaction with a single point. Additionally, ClickVOS offers the flexibility for algorithms to concentrate on parts. In ClickVOS, it's possible to flexibly select objects by clicking on background points to suppress the foreground (points indicating other objects can also be considered as background points for the current object).

%% file: sections/method.tex
\section{Attention Before Segmentation}

\begin{figure*}[t]
    \centering
    \includegraphics[width=1.0\textwidth]{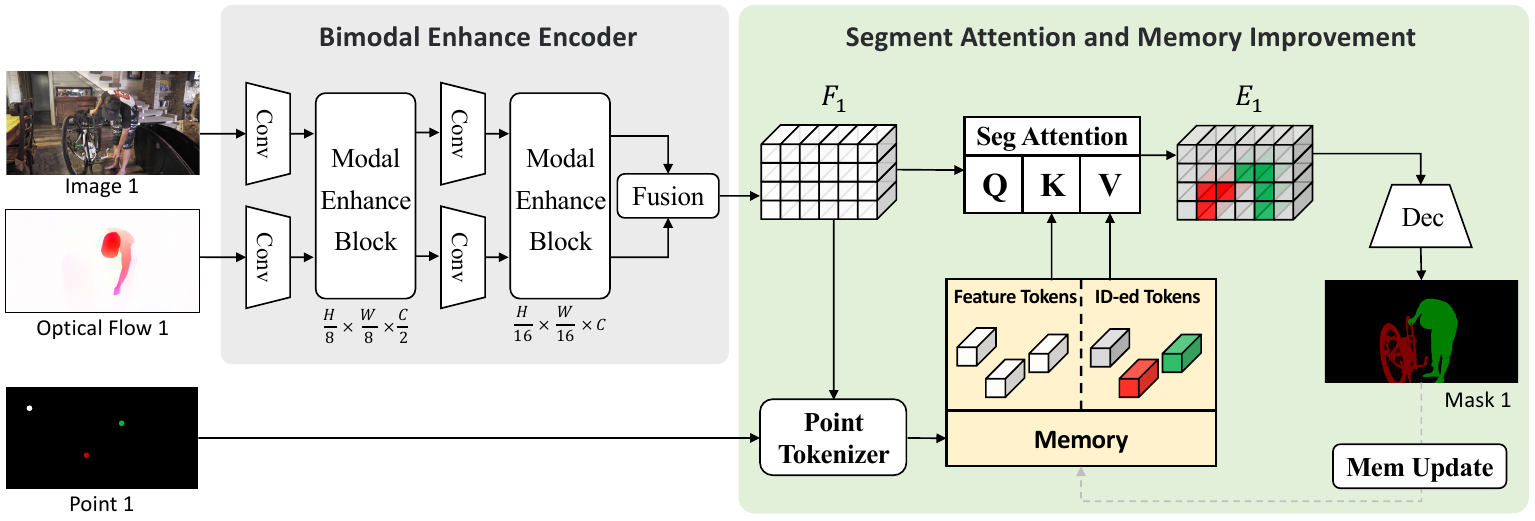}
    \caption{The pipeline of the proposed ABS approach for the ClickVOS problem. We extract bimodal features that contain appearance and motion information by the Bimodal Enhance Encoder. At the first frame, Point Tokenizer encodes the object tokens with identity embedding according to the given points, and Segment Attention estimates the initial object masks, which may be imprecise. But as the video progresses, a growing memory is maintained, leading to the self-healing of object masks.}
    \label{fig_overview}
\end{figure*}

We propose an end-to-end approach, Attention Before Segmentation (ABS), to tackle the ClickVOS task. Fig. \ref{fig_overview} shows the overview of our method. First, we extracted the bimodal feature incorporating appearance and motion information with the Bimodal Enhance Encoder. For the first frame, tokens with identity embedding for each object were obtained with the Point Tokenizer with the given points. The initial object masks are estimated through the Segment Attention with the object tokens and the bimodal feature. These object masks may still be imprecise with some errors. As the video goes on, a growing memory is built up and maintained, which makes object masks can be increasingly precise.

\subsection{Bimodal Enhance Encoder}
\label{bimodal_encoder}

The annotation point at the first frame could provide the position of the object. However, obtaining an accurate segmentation mask for the object solely based on its approximate position in the image is still a challenging problem. Fortunately, for objects in the video, motion information can be of great benefit in predicting the shape of objects. To achieve this, we utilize both the appearance information from the RGB image and the motion information from the optical flow. We propose the Bimodal Enhance Encoder, which could deeply couple the information from both modalities, complementing each other to obtain enhanced bimodal features.

Given a video with $T$ frames $ \{I_t \in \mathbb{R}^{3\times H\times W}\}_{t=1}^{T} $, the optical flows $ \{O_t \in \mathbb{R}^{3\times H\times W}\}_{t=1}^{T} $ (visualized as RGB images) are calculated by RAFT \cite{teed2020raft}: $ O_t = \mathrm{RAFT}(I_{t-1}, I_t) $, and the optical flow of the first frame is multiplexed from the following frame: $ O_1=O_2=\mathrm{RAFT}(I_1, I_2) $. Residual convolution blocks from ResNet-50 \cite{resnet} are employed to extract features, and the modal enhance blocks are designed to merge the bimodal features. These two operations are interleaved multiple times in a repeated manner.

Specifically, the RGB image and optical flow are fed through the residual convolution blocks, resulting in their respective features: $f_t^{I_8}, f_t^{O_8} \in \mathbb{R}^{\tfrac{H}{8} \times \tfrac{W}{8} \times \tfrac{C}{2}}$.
To merge the features of two modalities, we utilize the multi-head attention~\cite{vaswani2017attention} to design the modal enhance block.
The features are first augmented by combining self-modality features from all other positions in a global context with self-attention:
\begin{equation}
    \label{self-att-I}
    f_t^{I_{8}^\prime} = \mathrm{Attn}(f_t^{I_8}, f_t^{I_8}, f_t^{I_8}) \in \mathbb{R}^{\tfrac{H}{8} \times \tfrac{W}{8} \times \tfrac{C}{2}}
\end{equation}
\begin{equation}
    \label{self-att-O}
    f_t^{O_{8}^\prime} = \mathrm{Attn}(f_t^{O_8}, f_t^{O_8}, f_t^{O_8}) \in \mathbb{R}^{\tfrac{H}{8} \times \tfrac{W}{8} \times \tfrac{C}{2}}
\end{equation}
where $\operatorname{Attn}(\mathbf{Q}, \mathbf{K}, \mathbf{V})=\operatorname{softmax}\left(\frac{\mathbf{Q} \mathbf{K}^{\top}}{\sqrt{d_k}}\right) \mathbf{V}$, and $ d_k $ is the key dimensionality. For brevity, the projection operations of Q, K, and V in the attention are omitted here. 
Next, cross-attention is performed between the bimodel features, allowing them to acquire complementary information from the other modality.
\begin{equation}
    \label{cross-att-I}
    f_t^{IO_8} = f_t^{I_{8}^\prime} + \mathrm{Attn}(f_t^{I_{8}^\prime}, f_t^{O_{8}^\prime}, f_t^{O_{8}^\prime}) \in \mathbb{R}^{\tfrac{H}{8} \times \tfrac{W}{8} \times \tfrac{C}{2}}
\end{equation}
\begin{equation}
    \label{cross-att-O}
    f_t^{OI_8} = f_t^{O_{8}^\prime} + \mathrm{Attn}(f_t^{O_{8}^\prime}, f_t^{I_{8}^\prime}, f_t^{I_{8}^\prime}) \in \mathbb{R}^{\tfrac{H}{8} \times \tfrac{W}{8} \times \tfrac{C}{2}}
\end{equation}

By repeatedly applying the residual convolutional and bimodal enhance block as above, we obtain mutually enhanced bimodal features $ f_t^{IO_{16}}, f_t^{OI_{16}} \in \mathbb{R}^{\tfrac{H}{16} \times \tfrac{W}{16} \times C} $. Finally, a channel-attention module is employed to fuse the enhanced bimodal features, resulting in the final bimodal feature.
\begin{equation}
    \label{fusion}
    F_t = \mathrm{Fusion}( [f_t^{IO_{16}}, f_t^{OI_{16}}] ) \in \mathbb{R}^{\tfrac{H}{16} \times \tfrac{W}{16} \times C}
\end{equation}

\subsection{Objects Perception with Point}
\label{seg_attention}

\begin{figure*}[t]
    \centering
    \includegraphics[width=0.98\textwidth]{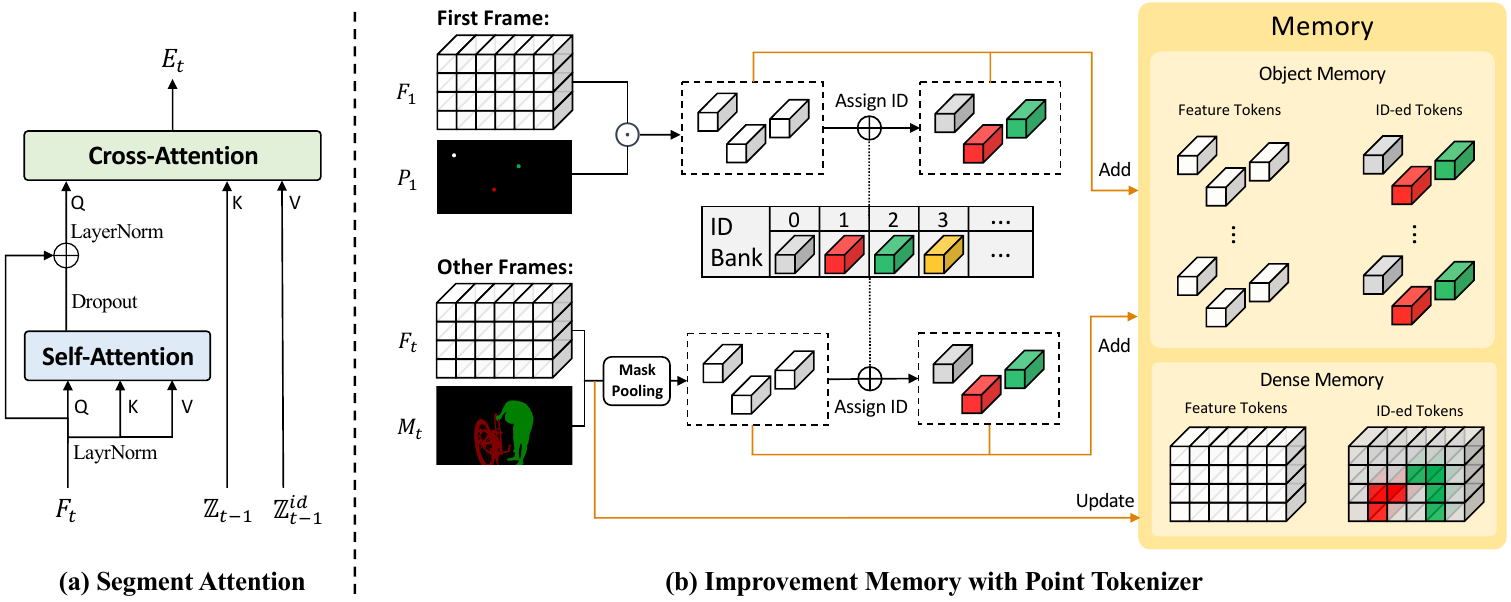}
    \caption{Details of the (a)~Segment Attention that achieve segmentation by simple attention layers, and the (b)~Improvement Memory with Point Tokenizer.}
    \label{fig_module}
\end{figure*}

With the bimodal feature enhanced by motion information, we can employ the designed Segment Attention module and given points to segment the interested objects. As illustrated in Fig.~\ref{fig_module}~(b), bimodal feature are first sampled into the object feature tokens $ z_1 \in \mathbb{R}^{N \times C} $ according to the object points of the first frame. These feature tokens are then encoded into object ID-ed tokens $ z_1^{id} \in \mathbb{R}^{N \times C} $ by adding the identity embeddings $ id \in \mathbb{R}^{N \times C} $ selected from the ID bank:
\begin{equation}
    \label{ided}
    z_1^{id} = z_1 + id
\end{equation}
where identity embeddings are learnable parameters that indicate which object the feature corresponds to. The identity mechanism, which associates multiple objects into the same high-dimensional embedding space proposed in \cite{aot}, allows our Segment Attention to handle the matching and segmentation of multiple objects simultaneously. With our concise yet functional Segment Attention module, consisting of self-attention and cross-attention as depicted in Fig.~\ref{fig_module}~(a), the dense ID-ed embedding $ E_1 \in \mathbb{R}^{\tfrac{H}{16} \times \tfrac{W}{16} \times C} $ of the first frame with object identity embeddings are encoded,
\begin{equation}
    \label{seg_attn}
    E_1 = \mathrm{SegAttn}(F_1, z_1, z_1^{id})
\end{equation}
which can be decoded into the objects mask $ M_1 \in \mathbb{N}^{H\times W} $ with a convolutional decoder made up of a series of residual blocks:
\begin{equation}
    \label{dec}
    M_1 = \mathrm{Dec}([E_1, F_1]) \in \mathbb{N}^{H\times W}
\end{equation}
where $\operatorname{Dec}$ indicates the decoder and $[~,~]$ refers to the concat operation.

\subsection{Masks Self-healing with Improvement Memory}
\label{mem_improvement}

Upon prediction of the object mask in the first frame by the segment attention mechanism, the segmentation details of the object can be gradually improved as the video sequence goes on. This is made possible by our proposed Improvement Memory, which comprises object memory and dense memory. 
Specifically, upon the mask of a frame is predicted, the object tokens of the object are computed and appended to the object memory, while the dense tokens of the frame are calculated and updated into the dense memory. As shown in Fig. \ref{fig_module}~(b), the bimodal feature of the current frame is aggregated into the object feature tokens $ z_t $ by the mask pooling with the predicted mask:
\begin{equation}
    \label{pool}
    z_t = \mathrm{MaskPooling}(F_t,M_t) = \frac{\sum_{i,j} F_t^{i,j} \cdot \overline{M}_t^{i,j}}{\sum_{i,j} \overline{M}_t^{i,j}}
\end{equation}
where $ \overline{M_t} \in \mathbbm{1}^{N \times \tfrac{H}{16} \times \tfrac{W}{16}} $ is the binary mask after downsampling and one-hot encoding. The indices $i$ and $j$ represent the position of the feature and binary mask. Then, the identity embeddings from the ID bank are added to obtain object ID-ed tokens, following the same procedure as in the first frame: $ z_t^{id} = z_t +id $.
Moreover, we calculate a pair of dense tokens, which are retained in dense memory to provide detailed information about the object. Specifically, the bimodal feature of the current frame is directly utilized as the dense feature tokens, and the dense ID-ed tokens are obtained by adding the identity embeddings of the corresponding position from the ID-bank to the features based on the predicted mask.
\begin{equation}
    \label{dense tokens}
    Z_t = F_t, \  Z_t^{id} = F_t + id, \  \in \mathbb{R}^{H \times W \times C}
\end{equation}
The size of the dense memory is constrained to only two frames, encompassing the dense tokens of the previous frame and the first frame.

The object memory and dense memory jointly form the overall memory that retains crucial information regarding target objects. This memory is continually enhanced as the video goes on, with the addition of global object tokens and the update of detailed dense tokens, leading to a progressively precise segmentation of the objects.
All feature tokens stored in memory $ \mathbb{Z}_t = \{ z_t \}_1^t \cup \{ Z_1, Z_t \} \in \mathbb{R}^{ (tN+2HW) \times C} $ are utilized as the key, and all $\mathbb{Z}_t^{id} = \{ z_t^{id} \}_1^t \cup \{ Z_1^{id}, Z_t^{id} \} \in \mathbb{R}^{ (tN+2HW) \times C}$ ID-ed tokens stored in memory serve as the value for our Segment Attention mechanism. 

%% file: sections/baseline.tex
\section{Baseline Exploration}

We investigate utilizing off-the-shelf algorithms from related fields to address the ClickVOS task, including SemiVOS, UnVOS, InterVOS, instance segmentation, and interactive image segmentation. As a result of these explorations, multiple baselines are established for ClickVOS. These efforts can offer valuable insights and experiences for future research. Furthermore, comparative experiments with these baselines also serve to demonstrate the effectiveness of the proposed AttSeg approach.

    \textbf{BL-UnVOS}. A naive solution can be implemented utilizing UnVOS models capable of consistently segmenting potential foreground objects in each frame, followed by selecting the target objects by given points in the first frame. Despite most UnVOS methods only segment salient objects at the object level, making it difficult to predict multiple objects individually. Fortunately, RVOS~\cite{ventura2019rvos} can predict multiple salient objects simultaneously and individually. We utilize RVOS to predict multiple object masks at the instance level throughout the video and select the objects of interest with the given points in the first frame.
    
    \textbf{BL-InterVOS}. Another simple solution involves inputting the initial frame's click directly into an InterVOS methodology that supports the click interaction. We select MiVOS~\cite{mivos} here which integrates fBRS~\cite{fbrs2020} to obtain object masks of one frame and STM~\cite{oh2019stm} for propagating the masks to the entire video.

    \textbf{BL-SemiVOS}. A natural approach to tackling the ClickVOS problem robustly is transforming it into the more extensively studied SemiVOS problem if target masks in the first frame can be obtained from the given points. In order to establish the initial masks for target objects, one viable method is to leverage instance segmentation algorithms such as MaskRCNN~\cite{maskrcnn}. So we provide a two-stage baseline, BL-SemiVOS, where MaskRCNN predicts all instance masks in the first frame, and the object masks of interest are selected according to the given points. Finally, these object masks are propagated to each frame with the help of the SemiVOS method STM~\cite{oh2019stm}.
    
    \textbf{BL-SAMTrack}. Moreover, interactive image segmentation algorithms for data annotation also can achieve the points-to-mask process. Particularly, the recent advancements in SAM~\cite{kirillov2023segment}, a foundation model for segmentation, have yielded exciting results. We utilize clicks as prompts for SAM, so SAM can generate the initial target masks. Subsequently, a state-of-the-art SemiVOS algorithm DeAOT~\cite{yang2022decoupling} is employed to propagate the object masks, resulting in a strong baseline termed BT-SAM-Track. Notably, this pipeline is implemented within SAM-Track~\cite{cheng2023segment} and we can reuse the implementation to address the  ClickVOS directly by providing only a single click per object in the first frame.

    \textbf{BL-SAM-PT}. Based on the powerful SAM, SAM-PT~\cite{rajivc2023segment} introduces an alternative approach for interactively segmenting objects in videos which can serve as another SAM-based baseline, BL-SAM-PT. Given the clicks in the initial frame, BL-SAM-PT utilizes a point tracker to predict corresponding points in each subsequent frame. These points are then utilized as prompts of SAM, resulting in segmentation masks for the objects in each frame.

\begin{figure}[t]
    \centering
    \includegraphics[width=1.0\linewidth]{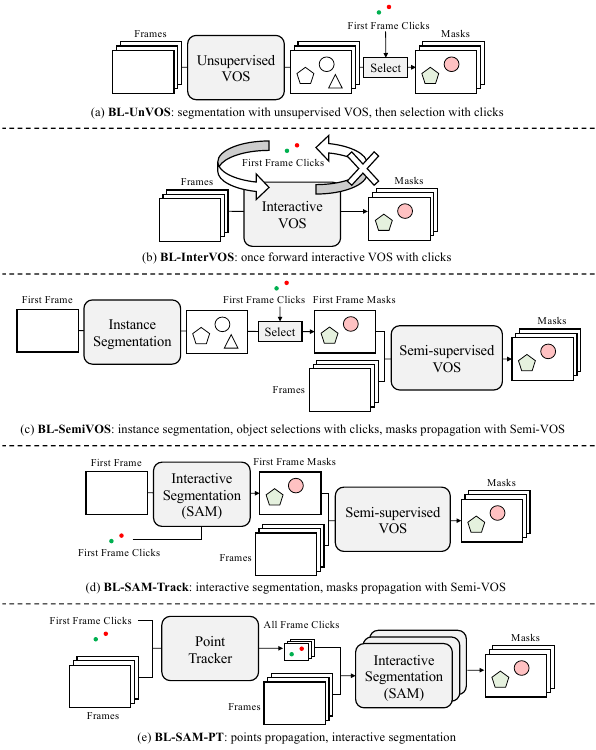}
    \caption{Illustration of baselines exploration of utilizing off-the-shelf algorithms from related fields to address the ClickVOS task.}
    \label{fig_baselines}
\end{figure}

%% file: sections/experiments.tex
\section{Experiments}

\begin{table*}[]
    \caption{Click video object segmentation performance comparisons between the proposed approach and the established baselines on the validation sets of DAVIS-2016, DAVIS-2017, and YouTubeVOS-2019.}
    \centering
    \begin{tabular}{ccccclccclccc}
        \toprule
          & & \multicolumn{3}{c}{DAVIS-2016} &  & \multicolumn{3}{c}{DAVIS-2017} &  & \multicolumn{3}{c}{YouTubeVOS-2019} \\
        Approach  & Strategy & \(\mathcal{J\&F}\) & \(\mathcal{J}\) & \(\mathcal{F}\) &  & \(\mathcal{J\&F}\) & \(\mathcal{J}\) & \(\mathcal{F}\) &  & \(\mathcal{J\&F}\) & \(\mathcal{J}\) & \(\mathcal{F}\) \\
        \midrule
        BL-UnVOS~\cite{ventura2019rvos} & UnVOS + Point Selection & 64.8 & 64.8 & 64.8 &  & 41.2 & 48.0 & 52.6 &  & 33.7 & 33.0 & 34.5 \\
        BL-InterVOS~\cite{mivos}  & InterVOS with Point & 70.5 & 70.0 & 71.0 &  & 63.5 & 60.0 & 67.1 &  & 40.9 & 38.7 & 43.4  \\
        BL-SemiVOS~\cite{maskrcnn,oh2019stm} & InstanceSeg + MaskTracking & 75.0 & 74.7 & 75.5 &  & 64.6 & 62.6 & 66.7 &  & 38.6 & 37.7 & 39.5 \\  
         BL-SAM-Track~\cite{cheng2023segment}  & PointSeg + MaskTracking & 73.8 & 74.1 & 73.6 &  & 66.3 & 63.3 & 69.3 &  & 42.9 & 40.8 & 45.3 \\
        BL-SAM-PT~\cite{rajivc2023segment}  & PointTracking + PointSeg & 67.0 & 66.9 & 67.1 &  & 58.5 & 55.8 & 61.3 &  & 44.2 & 41.9 & 46.6 \\
        
        ABS (ours) & End-to-end & \textbf{85.0} & \textbf{85.2} & \textbf{84.9} &  & \textbf{67.6} & \textbf{64.2} & \textbf{70.9} &  & \textbf{46.9} & \textbf{45.4} & \textbf{48.4} \\
        \bottomrule
    \end{tabular}
    \label{tab_compare}
\end{table*}

\begin{figure*}[t]
    \centering
    \includegraphics[width=1.0\linewidth]{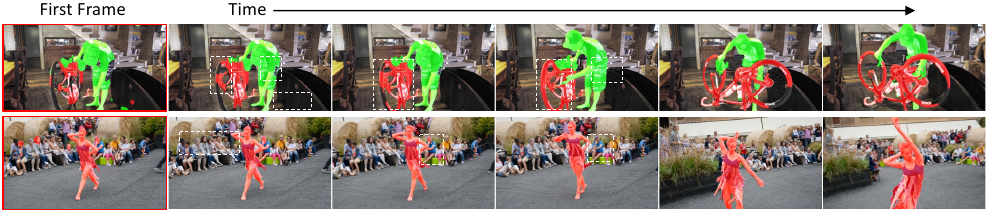}
    \caption{The self-healing process in our proposed ABS approach,
    White dashed boxes shows segmentation self-healing as the video goes on.}
    \label{fig_improve}
\end{figure*}

\begin{figure*}[]
    \centering
    \includegraphics[width=1.0\linewidth]{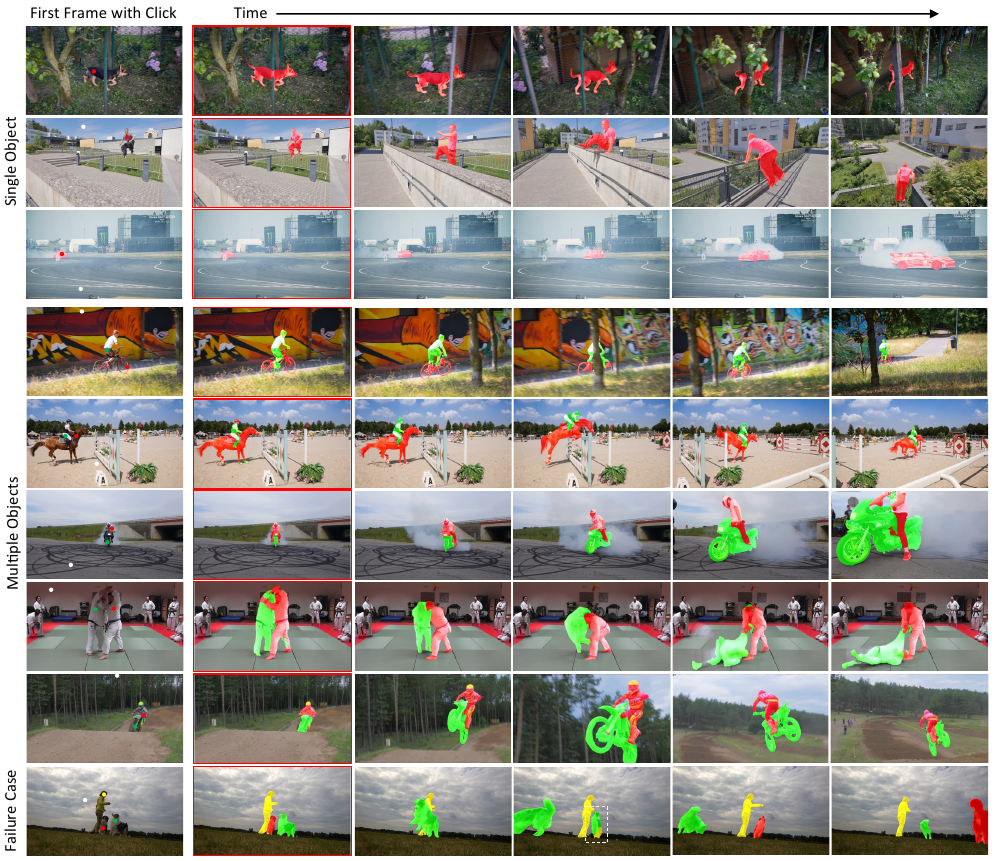}
    \caption{Qualitative results of our approach. The first column shows the first frame with the click points. Red outer borders indicate the first frame. The last row shows a failure case as the white dashed box.
    }
    \label{fig_results}
\end{figure*}

\begin{figure*}[t]
    \centering
    \includegraphics[width=1.0\linewidth]{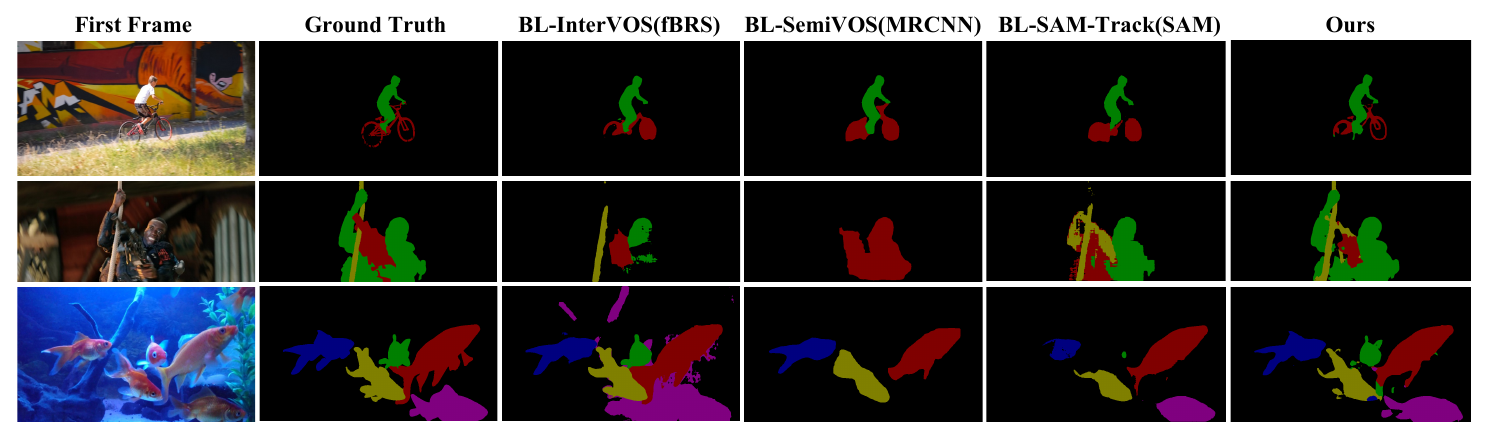}
    \caption{Qualitative results at the first frame of our method in comparison with other established baselines using the existing algorithms from related fields.}
    \label{fig_f0}
\end{figure*}

\subsection{Implementation Details}
We adapt the ResNet-50 \cite{resnet} pre-trained on ImageNet \cite{imagenet} to serve as the convolutional blocks in our bimodal feature encoder. Specifically, we utilize the blocks before conv3 as the first convolutional block and the conv4 block as the second convolutional block. 
Our decoder is composed of a sequence of residual convolutional blocks, with each block progressively transforming and upsampling features by a factor of two.
In Segment Attention, the channel dimension is 1024 and the head number is set to 8. We use the standard multi-head attention implemented in PyTorch for global object memory and the long-short attention implemented by \cite{aot} for detailed dense memory.
The proposed model is trained on the DAVIS-2017~\cite{davis17} and YouTubeVOS \cite{ytvos18} datasets with four NVIDIA GeForce RTX 3090 GPUs. We train the model with the training samples by simulating the inference phase. Specifically, the model is supposed to predict the mask of the first frame with the given points and to predict subsequent frame masks using the previous frames and their predicted masks. The model is trained for 150k iterations, with a batch size of 8, minimizing the bootstrapped cross-entropy loss and dice loss with equal weighting. We utilize the Adam optimizer \cite{kingma2014adam} with an initial learning rate of $10^{-5}$, which drops to $10^{-6}$ after 120k iterations.

\subsection{Evaluation and Comparision}

As illustrated in Table \ref{tab_compare}, we evaluate our approach and compare with the above baselines on the DAVIS-2016, DAVIS-2017, and YouTubeVOS-2019 validation set. Our proposed ABS shows superior performance compared to all baselines on all three datasets, even though some models have been trained on additional datasets. On the single-object benchmark DAVIS-2016, ABS surpasses the best-performing baseline by a significant margin. On the multiple-object benchmark DAVIS-2017 and YouTubeVOS, our method also achieves the best accuracy among all competing baselines. It was observed that for unseen objects in YouTubeVOS, MiVOS performs better. This is attributed to the fact that the interactive segmentation model fBRS in MiVOS is trained with additional datasets, including COCO \cite{lin2014coco} with 80 categories and LVIS \cite{gupta2019lvis} with 1203 categories, resulting in many unseen object classes of YouTubeVOS that have been trained to be no longer unseen. 

Fig. \ref{fig_results} shows the qualitative results of the proposed method.  Our method not only segments the single object well but also can handle multiple objects across the entire video. The last row also shows a failure case where the video contains a person and two similar dogs with frequent interaction and occlusion. The person was always segmented accurately, but because one dog completely occluded the other for a period of time, the mask of the occluded dog was incorrectly predicted when it reappeared in the frame. However, thanks to our self-improving memory mechanism, the error can be corrected in subsequent frames. 

\begin{table}[]
    \setlength\tabcolsep{3pt}
    \caption{Ablation study on the DAVIS-2017 validation set.}
    \centering
      \begin{tabular}{cccc|ccc}
      \toprule
      ~ & \footnotesize Seg Attention & \footnotesize Bimodal & \footnotesize Improvement & \multirow{2}*{\(\mathcal{J\&F}\)} & \multirow{2}*{\(\mathcal{J}\)} & \multirow{2}*{\(\mathcal{F}\)}\\
      ~ & \footnotesize with Point  & \footnotesize Enhance & \footnotesize Memory & ~ & ~ & ~ \\
  
      \midrule
      \footnotesize \textcolor{darkF7E0D5}{1} & \checkmark & & & 44.6 & 41.7 & 47.6\\
      \footnotesize \textcolor{darkF7E0D5}{2} & \checkmark &  & \checkmark & 56.2 & 52.8 & 59.7\\
      \footnotesize \textcolor{darkF7E0D5}{3} & \checkmark & \checkmark & & 54.8 & 51.2 & 58.4\\
      \footnotesize \textcolor{darkF7E0D5}{4} & \checkmark & \checkmark & \checkmark & \textbf{67.6} & \textbf{64.2} & \textbf{70.9} \\
  
      \bottomrule
      \end{tabular}
    \label{tab_ablation}
\end{table}

\subsection{Ablation Study}

In this section, we analyze the main components of our model and evaluate their impact. All ablation studies are conducted on the DAVIS-2017 benchmark.

The line \textcolor{darkF7E0D5}{1} of Table \ref{tab_ablation} indicates the simplest model capable of performing ClickVOS with our Segment Attention module, where only RGB image is used for encoding the feature, and the memory only stores the object tokens obtained from the given points in the first frame. Our Bimodal Enhance Encoder, which fuses and enhances features from both appearance and motion modalities, improves the performance from 44.6 to 54.8 $\mathcal{J\&F}$ without improvement memory (Line \textcolor{darkF7E0D5}{3}) and further enhances the performance from 56.2 to 67.6 $\mathcal{J\&F}$ with the improvement memory (Line \textcolor{darkF7E0D5}{4}). By designing the improvement memory, our model can self-correct incorrect segmentations of objects in subsequent frames, leading to an overall improvement in performance from 44.6 to 56.2 $\mathcal{J\&F}$ (Line \textcolor{darkF7E0D5}{2}) and 54.8 to 67.6 $\mathcal{J\&F}$ (Line \textcolor{darkF7E0D5}{4}).

\subsubsection{Ability of Objects Perception}

To address the challenge of perceiving target objects according to the given points in the first frame within the ClickVOS task, ABS queries all pixels in the first frame with ID-ed object tokens through our segmentation attention. 
Fig.~\ref{fig_f0} illustrates our approach's superior performance compared to other baselines, including the interactive image segmentation model fBRS, instance segmentation model MRCNN, and even the foundation model SAM.     
To quantitatively evaluate the proposed SegAttn perceiving objects in the first frame by the simple attention, we compare the segmentation quality in the first frame with other segmentation models employed in baselines.
Specifically, the BL-InterVOS approach adopts fBRS, BL-SemiVOS utilizes MaskRCNN, and BL-SAM-Track and BL-SAM-PT employ the foundational models SAM-Base and SAM-Huge, respectively. As illustrated in Fig.~\ref{fig_f0} and Tab.~\ref{tab_f0_comparison}, in the case where a point can only provide location information, our SegAttn simply and effectively accomplishing object perception is even not inferior to other baselines that employ interactive segmentation models, even in challenging scenarios such as background interference and multiple similar objects. Moreover, our approach won't lose target objects, compared with BL-SemiVOS which employs the instance segmentation model MaskRCNN to predict all possible objects first, followed by the mask selection with points.

\begin{table}[t]
\centering
\caption{Comparison of objects perception in the first frame.}
\begin{tabular}{l|ccc}
\toprule
\multicolumn{1}{c|}{Method} &\(\mathcal{J\&F}\)&\(\mathcal{J}\)&\(\mathcal{F}\)\\
\midrule
BL-InterVOS                & 73.6 & 68.8 & 78.5 \\
BL-SemiVOS                 & 66.1 & 63.7 & 68.6 \\
BL-SAM-Track               & 66.5 & 62.3 & 70.6 \\
BL-SAM-PT                  & 78.9 & 76.1 & 81.8 \\
Ours                       & 73.8 & 69.9 & 77.7 \\
\bottomrule
\end{tabular}
\label{tab_f0_comparison}
\end{table}

\subsubsection{Ability of Self-healing}

\begin{table}[t]
    \caption{Ablation study of the improvement memory. 
    ObjMem$\times 1$ represents only the object tokens of the first frame used in the whole video. ObjMem$\times T$ indicates that object tokens of each previous frame are added to the memory.
    }
    \centering
        \begin{tabular}{l|ccc}
        \toprule
        \multicolumn{1}{c|}{Memory} & \(\mathcal{J\&F}\) & \(\mathcal{J}\) & \(\mathcal{F}\)\\
        \midrule
        ObjMem$\times 1$            & 54.8 & 51.2                   & 58.4 \\
        ObjMem$\times T$            & 63.6 & 60.0                   & 67.1 \\
        ObjMem$\times 1$ + DenseMem & 56.4 & 52.6                   & 60.3 \\
        ObjMem$\times T$ + DenseMem & \textbf{67.6} & \textbf{64.2} & \textbf{70.9} \\ 
        \bottomrule
        \end{tabular}
    \label{tab_ablation_mem}
\end{table}

As illustrated in Fig. \ref{fig_improve}, by employing our designed improvement memory, even if the object mask of the first frame is predicted imprecisely, masks of subsequent frames can perform self-improvement as the video evolves instead of occurring the error accumulation. Because the object tokens in memory store the overall feature increasingly, and the dense tokens in memory contain pixel-level details that are updated frame by frame. The ablation study on improvement memory is shown in Table \ref{tab_ablation_mem}, which demonstrates that growing object tokens can largely correct predictions in subsequent frames and that dense tokens can improve the detail of segmentation masks.

\subsubsection{Effectiveness of Bimodal Enhance Encoder}

\begin{table}[t]
    \caption{Ablation study of the bimodal enhance encoder. Appearance indicates only extracting the RGB image feature with a ResNet-50. Appearance+Motionusing represents the appearance fusing the appearance and motion features extracted by the ResNet-50 with a simple fully connected layer. Bimodal Encoder is the Bimodal Enhance Encoder proposed in this paper.}
    \centering
    \begin{tabular}{lccc}
    \toprule
    \multicolumn{1}{c|}{Modality}       & \(\mathcal{J\&F}\) & \(\mathcal{J}\) & \(\mathcal{F}\)\\ 
    \midrule
    \multicolumn{4}{c}{\small \textit{Segment with only ObjMem$\times 1$} }                                  \\
    \hline
    \multicolumn{1}{l|}{Appearance}   & 44.6          & 41.7          & 47.6          \\
    \multicolumn{1}{l|}{Appearance + Motion} & 46.9          & 44.1          & 49.8          \\
    \multicolumn{1}{l|}{Bimodal Encoder}   & \textbf{54.8} & \textbf{51.2} & \textbf{58.4} \\ 
    \midrule
    \multicolumn{4}{c}{\small \textit{Segment with full Mem} }                                     \\
    \hline
    \multicolumn{1}{l|}{Appearance}   & 56.2          & 52.8          & 59.7          \\
    \multicolumn{1}{l|}{Appearance + Motion} & 63.7          & 60.1          & 67.2          \\
    \multicolumn{1}{l|}{Bimodal Encoder}   & \textbf{67.6} & \textbf{64.2} & \textbf{70.9} \\ 
    \bottomrule
    \end{tabular}
    \label{tab_ablation_modal}
\end{table}

As shown in Table \ref{tab_ablation_modal}, we analyze the effect of our bimodal enhance encoder. When segmenting objects with only the object tokens of the first frame, fusing the appearance and motion information with a simple fully connected layer marginally improves object segmentation performance. Fusing and enhancing bimodel feature with our Bimodal Enhance Encoder results in a significant boost in segmentation ability, as the deeply enhanced features with the motion modality could provide additional cues for the truly challenging problem that the whole object segmentation with only a position point. Furthermore, the experimental results obtained when the full memory is available further confirm this conclusion.

\subsubsection{Point Annotation}

We evaluate our model with different points of the first frame (manually annotated or automatically generated) in Table \ref{tab_ablation_point}. Overall, our method does not rely on specific points containing the human prior, and it can perform just as well with random points automatically selected from object masks. Manual clicks work slightly better in the multi-object benchmark, and even random points achieved higher performance for the single-object benchmark. These results also highlight the reasonableness of the automatically selected point annotation we provided for the training phase, which not only can reduce the need for human labour but also can ensure that the trained model is robust to points and won't rely on points with specific prior.

\begin{table}[t]
    \centering
    \caption{Ablation study on click points used in inference.}
    \begin{tabular}{c|ccc}
    \toprule
    \multirow{2}{*}{Point Annotation} & \multirow{2}{*}{\begin{tabular}[c]{@{}c@{}}D16\\ \(\mathcal{J\&F}\)\end{tabular}} & \multirow{2}{*}{\begin{tabular}[c]{@{}c@{}}D17\\ \(\mathcal{J\&F}\)\end{tabular}} & \multirow{2}{*}{\begin{tabular}[c]{@{}c@{}}Y19\\ \(\mathcal{J\&F}\)\end{tabular}} \\
                                      &                                                                                   &                                                                                   &                                                                                 \\
    \midrule
    Manual Click              & 85.0 & \textbf{67.6} & \textbf{43.9}\\
    Automatical from Mask    & \textbf{86.7} & 64.2 & 40.8\\ 
    \bottomrule
    \end{tabular}
    \label{tab_ablation_point}
\end{table}

\subsubsection{Tokens Selection in SAM-PT and SAM-track}
\begin{table}[t]
\caption{Mask token selection in SAM-based baselines.}
\centering
\begin{tabular}{c|c|ccc}
\toprule
Method                        & Mask Token  &\(\mathcal{J\&F}\)&\(\mathcal{J}\)&\(\mathcal{F}\)\\
\midrule
\multirow{4}{*}{BL-SAM-PT}    & unambiguous & 30.3 & 28.1 & 32.4 \\
                              & subpart     & 21.5 & 18.4 & 24.7 \\
                              & part        & 40.5 & 37.5 & 43.4 \\
                              & whole       & 58.5 & 55.8 & 61.3 \\
\midrule
\multirow{5}{*}{BL-SAM-Track} & confident   & 53.8 & 51.4 & 56.3 \\
                              & unambiguous & 49.1 & 46.4 & 51.9 \\
                              & subpart     & 40.2 & 37.8 & 42.5 \\
                              & part        & 49.8 & 46.9 & 53.7 \\
                              & whole       & 66.3 & 63.3 & 69.3 \\ 
\bottomrule
\end{tabular}
\label{tab_ablation_sam_token}
\end{table}

In baseline BL-SAM-PT and BL-SAM-Track, the segmentation model SAM is employed to segment the object mask. And in order to adapt to the different granularity of the target, SAM generates four tokens for each point prompt, which tend to segment the default region (unambiguous token), the whole object (whole token), the subpart (whole token), and the subpart (subpart token). We evaluated the performance of ClickVOS when using different tokens. The BL-SAM-Track implementation also provides an additional option: dynamically selecting the token with the higher SAM-predicted IoU Score. The experimental results show that using the token that tends to split out the whole object is the best in ClickVOS. 
So we choose the best performance they achieved when using whole-token to participate in the comparison experiment.

%% file: sections/conclusion.tex
\section{Conclusion}
We present a new task named ClickVOS, which aims at segmenting target objects in a video with only a single click per object in the first frame. And we provide the extended datatsets with points annotation to support this task. In addition, we propose an end-to-end approach ABS, which simulates the human attention process and achieves good performance for ClickVOS. 
Moreover, we conduct various baseline explorations utilizing off-the-shelf algorithms from related fields, which could provide experience for further research. And the experimental results demonstrate the superiority of the proposed ABS. We hope the new task and the above efforts will inspire the research community to explore new ideas and directions for video understanding.